\documentclass[letterpaper, 10 pt, conference]{ieeeconf}  

\IEEEoverridecommandlockouts                              

\usepackage{graphics} 
\usepackage{epsfig} 
\usepackage{times} 
\usepackage{amsmath} 
\usepackage{amssymb}  
\usepackage[backend=biber,sorting=nyt,sortcites=true]{biblatex}
\usepackage{multirow}
\usepackage{multicol}
\usepackage{hyperref}
\usepackage{booktabs}
\usepackage{enumerate}
\usepackage{color}
\usepackage{tabularx}
\usepackage{hyperref}
\usepackage{pifont}
\usepackage{algorithm}
\usepackage{algpseudocode}
\usepackage{amsmath}
\usepackage{adjustbox}
\usepackage[usenames,dvipsnames,table]{xcolor}
\hypersetup{
    colorlinks=true,
    linkcolor=MidnightBlue,
    filecolor=magenta,      
    urlcolor=MidnightBlue,
    citecolor=MidnightBlue,
} 
\usepackage[font=small,labelfont=bf]{caption}
\usepackage{marvosym}

\definecolor{darkred}{rgb}{0.76, 0.23, 0.13}
\definecolor{darkgreen}{rgb}{0.01, 0.75, 0.24}
\definecolor{darkgray}{rgb}{0.66, 0.66, 0.66}
\definecolor{customgray}{HTML}{F2F2F2}

\newcommand{\result}[2]{%
    \makebox[2.7em][r]{#1}%
    \hspace{0.2em}%
    \makebox[1.5em][l]{$^{\text{\color{darkgray}#2}}$}%
}

\newcommand{\resultgain}[2]{%
    \makebox[2.7em][r]{#1}%
    \hspace{0.2em}%
    \makebox[1.5em][l]{$^{\text{\color{darkred}#2}}$}%
}

\newcommand{\resultloss}[2]{%
    \makebox[2.7em][r]{#1}%
    \hspace{0.2em}%
    \makebox[1.5em][l]{$^{\text{\color{darkgreen}#2}}$}%
}

\title{\LARGE \bf
FP2: Equipping Robotic Foundation Models with Force Control
}

\author{
Hongjie Fang$^{1,2,4,*}$, Shirun Tang$^{1,2,*}$, Junjian Hu$^{1,2,5}$, Shidong Zhang$^{1,2,6}$, Derek Zhang$^{1,2,7}$, \\ Linhao Chen$^{1,2}$, Dehai Li$^{1,2}$, Mingyu Mei$^{8}$, Wanxi Liu$^{1,2,3}$, Cewu Lu$^{2,3,9}$, and Shiquan Wang$^{1,2,3,\text{\Letter}}$ 
\thanks{$^*$Equal Contribution. \space $^{\text{\Letter}}$Corresponding Author. \space $^1$FORTE Lab. \space $^2$Noematrix. \space $^3$Flexiv. \space $^4$SJTU. \space $^5$UPenn. \space $^6$FDU. \space $^7$UIUC. \space $^8$ZJU. \space $^9$SII.}
}

\begin{document}

\makeatletter
\let\@oldmaketitle\@maketitle
\renewcommand{\@maketitle}{
\@oldmaketitle
\vspace{0.2cm}
\centering
\includegraphics[width=\linewidth]{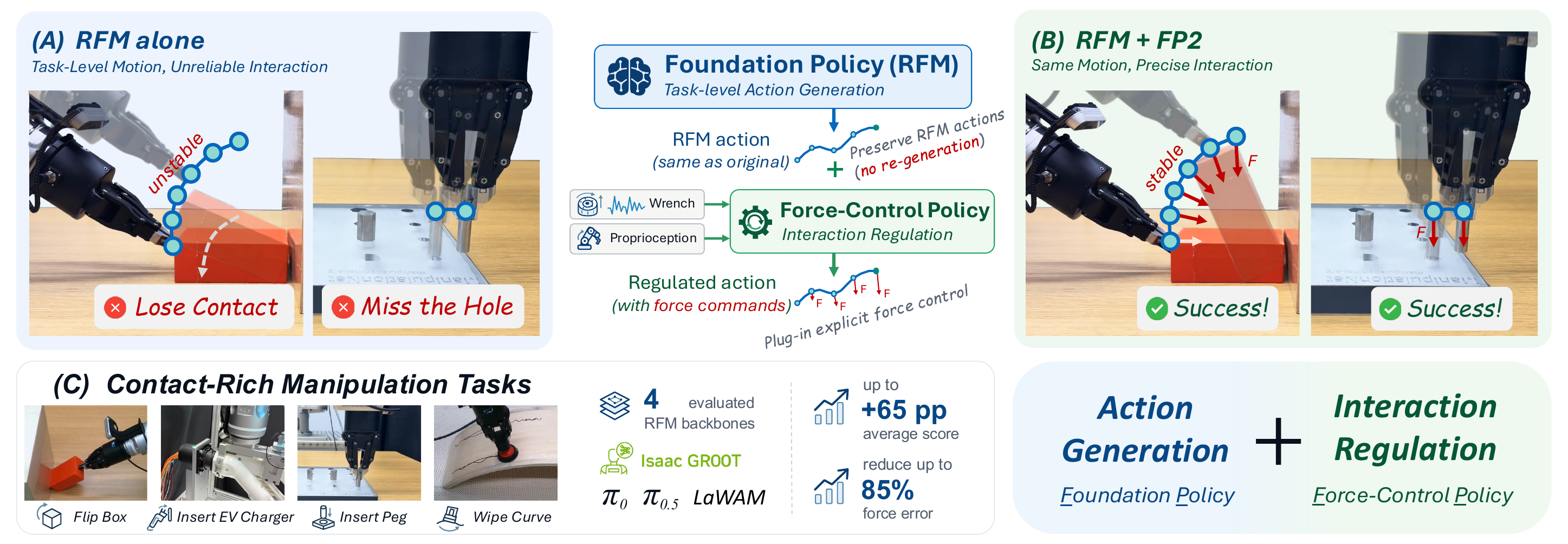}

\captionof{figure}{
\textbf{FP2: Decoupling Action Generation with Interaction Regulation.}
\textbf{\textit{(A)}} A task-adapted RFM can generate appropriate task-level motion yet still fail during physical interaction due to insufficient force regulation.
\textbf{\textit{(B)}} FP2 preserves foundation-policy action generation and adds a lightweight force control policy to regulate the same RFM actions using wrench and proprioceptive feedback, without action re-generation.
\textbf{\textit{(C)}} Across four RFM backbones and four contact-rich manipulation tasks, FP2 substantially improves task performance and force regulation, achieving up to 65 percentage-point gains in average score and reducing force error by up to 85\%.
}
\label{fig:teaser}\vspace{-0.3cm}
}%
\makeatother

\maketitle
\thispagestyle{empty}
\pagestyle{empty}
\addtocounter{figure}{-1}

\begin{abstract}
Robotic foundation models (RFMs) are increasingly capable of general-purpose manipulation, yet reliable physical interaction remains challenging in contact-rich settings. We present FP2, a lightweight downstream interface that equips task-adapted RFMs with explicit force control while preserving their action-generation capability. FP2 adopts an action-regulation decomposition: the task-adapted RFM serves as a foundation policy responsible for task-level action generation, while a high-frequency force control policy focuses solely on interaction regulation. To condition force regulation on the ongoing manipulation, FP2 compresses foundation-policy contextual representations and combines them with wrench and proprioceptive histories to predict structured force-control parameters. We evaluate FP2 with four RFM backbones across four real-world contact-rich manipulation tasks. FP2 consistently improves task performance and force regulation quality over the corresponding foundation policies, while comparing favorably with representative force-aware and force-control baselines. Ablations further show that foundation-policy context and physical feedback are complementary for effective force regulation, while preserving foundation-policy action generation improves both efficiency and novel-object generalization. Project website: \href{http://force-policy.github.io/fp2}{http://force-policy.github.io/fp2}
\end{abstract}

\section{Introduction}

Robotic foundation models (RFMs), including vision-language-action (VLA) models~\cite{openvla, octo, pi0, pi05, gr00t, sf} and world models~\cite{lingbotva, fastwam, cosmos}, have made rapid progress toward general manipulation by scaling visuomotor learning across diverse tasks and environments~\cite{oxe, rh20t, droid}. However, knowing what motion to execute does not necessarily imply knowing how to physically interact with the environment. In contact-rich manipulation, even a plausible task-level motion can fail under small uncertainties in geometry, friction, or compliance, leading to excessive forces, unstable contact, or jamming~\cite{suomalainen2022survey, tsuji2025survey}. As RFMs become increasingly capable at task-level action generation, robust physical interaction remains a critical bottleneck toward reliable real-world manipulation.

A natural approach is to integrate force feedback into the RFM itself, augmenting its observations with force for action prediction~\cite{forcevla, tavla, forcevla2, lift}. However, force is typically absent from large-scale RFM pretraining recipes. Introducing this new modality changes the pretrained observation interface and requires the foundation policy to learn its alignment with existing visuomotor representations from downstream data. This raises a key question: \emph{does equipping an RFM with force control require modifying the foundation model to incorporate force as an additional sensing modality?}

We take a different view: \textbf{force control does not need to be learned inside the robotic foundation model}. Once adapted to a downstream task, an RFM provides a \emph{foundation policy} that already generates task-level manipulation actions and rich contextual representations, while what remains missing is explicit regulation of physical interaction during execution. We therefore retain action generation in the foundation policy and introduce a separate force-control policy dedicated to interaction regulation. The force-control policy is conditioned on both physical feedback and the context of the foundation policy, allowing interaction regulation to account for the ongoing manipulation behavior.

Based on this insight, we introduce FP2, a lightweight framework that follows an \emph{action-regulation} decomposition: the foundation policy generates task-level actions, while a high-frequency force-control policy predicts only structured parameters for interaction regulation. Building on the force-control formulation of Force Policy~\cite{forcepolicy}, FP2 preserves foundation-policy action generation instead of duplicating it downstream. To efficiently provide task context to the force-control policy, FP2 self-supervisedly compresses the RFM contextual token sequence into a compact representation and combines it with wrench and proprioceptive histories. After task adaptation, the foundation policy remains fixed, and force information is introduced only when training the downstream force-control policy. This design preserves the RFM's task-level action-generation capability while adding explicit force regulation downstream.

We evaluate FP2 with four RFM backbones~\cite{pi0,pi05,gr00t,lawam} across four real-world contact-rich manipulation tasks. FP2 consistently improves task performance and generally reduces normalized force error relative to the corresponding foundation policies, while also comparing favorably with both task-specific force-control policies and force-aware VLAs. Ablations show that foundation-policy context and physical feedback are complementary for force regulation, and that preserving foundation-policy action generation improves efficiency and novel-object generalization over downstream action re-generation. Failure analysis further shows that FP2 primarily suppresses interaction-related failures, while errors in task-level motion generation remain outside its primary scope. Together, these results support separating task-level action generation from high-frequency interaction regulation as an effective interface for equipping existing RFMs with force control. All data, checkpoints, code, and evaluation videos will be made publicly available.
\section{Related Works}\label{sec:related}

\subsection{Robotic Foundation Models}

Early precursors to modern RFMs focused on learning general visual representations~\cite{r3m}, grounding language for robot decision making~\cite{saycan}, high-level planning~\cite{voxposer}, and extracting object affordances~\cite{vrb}. A subsequent line of work directly predicts robot actions from multimodal observations, with VLAs~\cite{rt1,rt2,bcz,octo,openvla,oxe} becoming a prominent paradigm. Recent VLAs~\cite{pi0,pi05,gr00t,pi07,sf} further improve generalization across tasks and environments by scaling pretraining on heterogeneous robot datasets~\cite{oxe,rh20t,droid}. More recently, world models~\cite{lingbotva,cosmos,fastwam,lawam} have emerged as a new paradigm, predicting future visual states and using them for action generation, often through inverse dynamics models. Their visual prediction objectives also facilitate learning from large-scale human videos and in-the-wild data~\cite{ego4d,egoverse,airexo} that lack robot action annotations.

Despite their different formulations, modern RFMs primarily target task understanding and action generation from visual, language, and proprioceptive observations, leaving physical interaction largely implicit in the predicted actions. FP2 addresses this complementary problem without changing the standard RFM task-adaptation interface: once task-adapted, the RFM serves as a foundation policy for action generation, while force regulation is introduced downstream.

\subsection{Force-Aware Policy Learning}

Recent works on contact-rich manipulation incorporate force sensing as an additional observation alongside vision and proprioception for action prediction~\cite{makingsensevisiontouch,vtt}. Early methods augment imitation policies~\cite{act,dp,rise} with multimodal fusion~\cite{manipforce,foar}, curriculum learning~\cite{factr}, or high-frequency force branches~\cite{implicitrdp,rdp}. More recently, force/torque observations have been incorporated into VLAs~\cite{forcevla,forcevla2,tavla,lift} and world models~\cite{fawam}, allowing large robotic policies to directly condition their predictions on physical interaction feedback. These approaches integrate force information into the manipulation policy itself, requiring the foundation policy to be adapted to the additional sensing modality. In contrast, FP2 keeps force feedback outside the RFM: a self-supervised compressor provides a compact foundation-policy context, which is combined with physical feedback only in the downstream force-control policy.

\subsection{Force Control for Contact-Rich Manipulation}

Force control provides a complementary route to contact-rich manipulation by explicitly regulating physical interaction rather than relying on action prediction alone. Learning-based methods have been integrated with impedance or admittance controllers to predict reference motions, interaction forces, or compliance parameters such as stiffness~\cite{acp,tacdiffusion,compact,dipcom,equicontact}. Several works~\cite{forcemimic,forcevla2,forcepolicy} build on hybrid force-position control~\cite{raibert_hybrid,mason_hybrid,bruyninckx1996specification,interaction_frame}, which decomposes task space into force- and position-controlled directions and provides a structured interface between learned policies and low-level interaction control. Force Policy~\cite{forcepolicy} combines task context from an upper-level visuomotor policy~\cite{rise2} with local visual and force feedback to predict interaction structure for contact-rich manipulation. FP2 follows this structured force-control formulation but adapts it to foundation policies with a clearer separation of responsibilities: the foundation policy generates task-level actions, while the force-control policy predicts only the parameters for interaction regulation.

A recent, concurrent work~\cite{huang2026unireflex} similarly augments frozen generative policies with a high-frequency reflex module conditioned on action-head latents and physical feedback, but the reflex module also re-generates reference actions alongside force-control quantities. FP2 instead (1) compresses the RFM contextual tokens into a compact representation and (2) preserves foundation-policy action generation, restricting the downstream policy to structured interaction regulation; we validate both design choices in \S\ref{sec:ablation}.
\section{Method}\label{sec:method}

\begin{figure*}
    \centering
    \includegraphics[width=\linewidth]{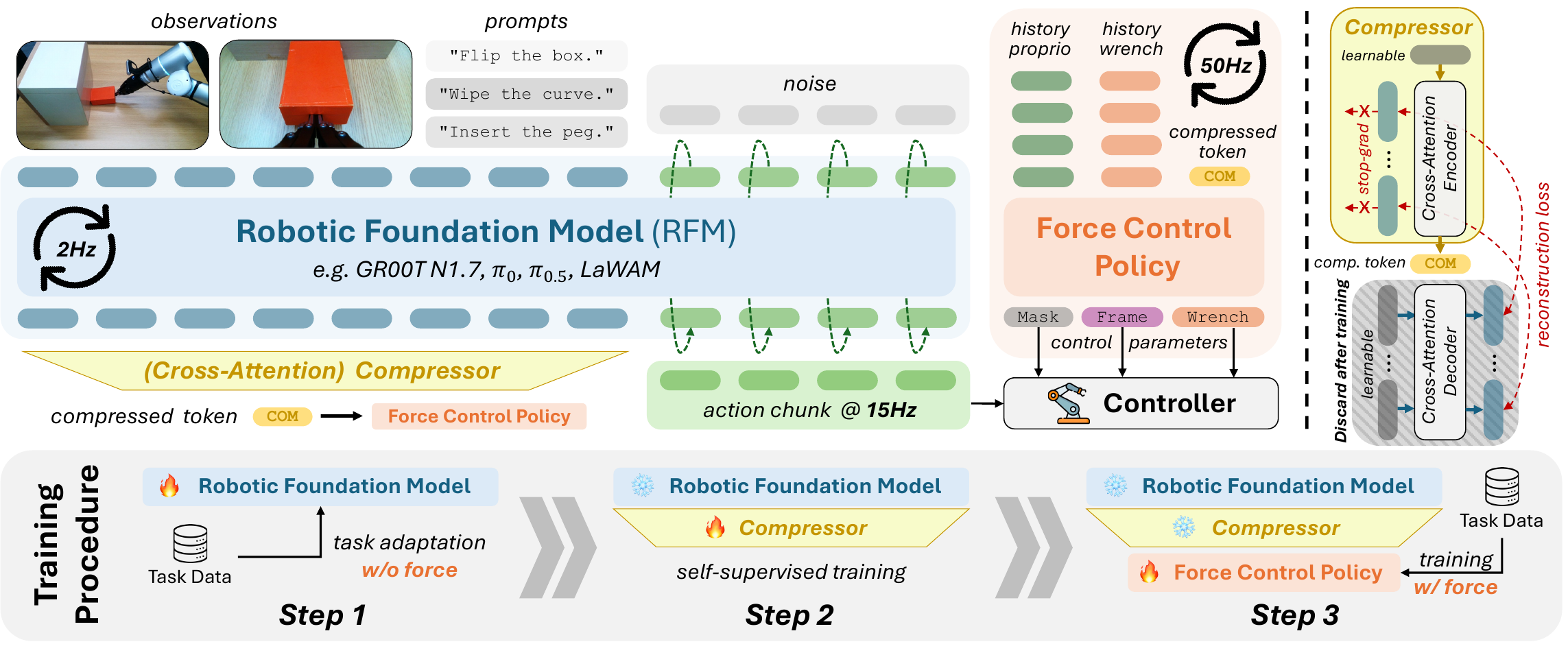}
    \caption{\textbf{FP2 Framework.}
    \textbf{\textit{(Top left)}} The task-adapted RFM serves as a foundation policy for action generation, while FP2 compresses its contextual tokens and combines them with wrench and proprioceptive histories to predict structured force-control parameters.
    \textbf{\textit{(Top right)}} The contextual token compressor is trained self-supervisedly through reconstruction; the reconstruction decoder is discarded after training.
    \textbf{\textit{(Bottom)}} Training proceeds in three stages: (1) standard RFM task adaptation, (2) self-supervised contextual token compression with the RFM frozen, and (3) force-control policy training with both the RFM and compressor frozen.}
    \label{fig:model}\vspace{-0.4cm}
\end{figure*}

\subsection{Overview and Problem Formulation}
\label{sec:overview}

As illustrated in Fig.~\ref{fig:model}, FP2 equips an RFM with force control without introducing force information into the RFM itself. We first adapt a pretrained RFM to downstream tasks following its standard task-adaptation procedure (Fig.~\ref{fig:model} (\textit{bottom}), \textbf{Step 1}), where the RFM uses only its native visuomotor interface and does not consume force measurements. After task adaptation, the RFM is frozen and serves as the \emph{foundation policy} $\pi$ for action generation. Let $\mathbf{a}_{t:t+H_a}=\pi(\mathbf{o}_t,\mathbf{l})$ denote a foundation policy action chunk~\cite{act} of horizon $H_a$ predicted from observation $\mathbf{o}_t$ and instruction $\mathbf{l}$.

FP2 augments the foundation policy with a contextual token compressor (\S\ref{sec:token}) and a lightweight force-control policy (\S\ref{sec:policy}). The compressor $\psi$ maps the high-dimensional contextual tokens $\mathbf{C}_t$ from $\pi$ into a compact token $\mathbf{z}_t=\psi(\mathbf{C}_t)$. The force-control policy $\pi_F$ combines $\mathbf{z}_t$ with recent wrench and proprioceptive history $\mathbf{W}_{t-H:t}, \mathbf{P}_{t-H:t}$ over a horizon $H$ to predict structured force-control parameters:
\begin{equation}
(\boldsymbol{\Sigma}_t,\mathbf{S}_t,\hat{\mathbf{W}}_t)
=
\pi_F
\left(
\mathbf{z}_t,
\mathbf{W}_{t-H:t},
\mathbf{P}_{t-H:t}
\right).
\label{eq:force_policy}
\end{equation}
Here, $\boldsymbol{\Sigma}_t$ denotes the interaction frame, $\mathbf{S}_t$ the force-position selection mask, and $\hat{\mathbf{W}}_t$ the desired wrench. These parameters, together with the foundation policy action $\mathbf{a}_t$, are passed to the robot controller (\S\ref{sec:control}) for execution.

Overall, the foundation policy determines the task-level motion, while the force-control policy specifies how that motion should be regulated under hybrid force-position control~\cite{raibert_hybrid, mason_hybrid,forcepolicy} during physical interaction.

\subsection{Contextual Token Compressor}
\label{sec:token}

RFM hidden representations typically contain many contextual tokens, making them inefficient to directly feed into a lightweight force-control policy running at high frequency. Following~\cite{rltoken}, we introduce a learned contextual-token bottleneck that compresses the token sequence into a compact representation (Fig.~\ref{fig:model} (\textit{bottom}), \textbf{Step 2}).

Given contextual tokens $\mathbf{C}_t\in\mathbb{R}^{S\times D}$, where $S$ denotes the sequence length and $D$ the hidden dimension, we first project each token into a $d$-dimensional embedding space. We then introduce a learnable query $\mathbf{Q}\in\mathbb{R}^{1\times d}$, which cross-attends to the projected tokens through a Transformer encoder $f_{\mathrm{enc}}$. Each encoder block consists of multi-head cross-attention followed by an MLP with residual connections. The resulting compressed token is
\begin{equation}
\mathbf{z}_t
=
\psi(\mathbf{C}_t)
=
f_{\mathrm{enc}}
\left(
\mathbf{Q},
\mathrm{Proj}(\mathbf{C}_t)
\right)
\in\mathbb{R}^{1\times d}.
\end{equation}

To train $\psi$ in a self-supervised manner, we additionally employ a lightweight Transformer reconstruction decoder~\cite{transformer}. As shown in Fig.~\ref{fig:model} (\textit{top right}), learned reconstruction queries attend only to $\mathbf{z}_t$ to reconstruct the original contextual tokens,
\begin{equation}
\hat{\mathbf{C}}_t
=
f_{\mathrm{dec}}(\mathbf{z}_t)
\in\mathbb{R}^{S\times D}.
\end{equation}
Thus, all information required for reconstruction must pass through the compressed bottleneck.

We optimize
\begin{equation}
\mathcal{L}
=
\mathcal{L}_{\mathrm{MSE}}
\left(
\mathbf{C}_t,\hat{\mathbf{C}}_t
\right)
+
\lambda
\mathcal{L}_{\mathrm{cos}}
\left(
\mathbf{C}_t,\hat{\mathbf{C}}_t
\right),
\end{equation}
where $\mathcal{L}_{\mathrm{MSE}}$ and $\mathcal{L}_{\mathrm{cos}}$ denote the token-wise MSE reconstruction loss and cosine-distance loss, respectively, and $\lambda$ controls their relative weight. After training, the decoder is discarded and only the compressor is retained for downstream force control.

During execution, $\mathbf{z}_t$ is updated synchronously with the foundation policy at 2\,Hz and cached between consecutive predictions. The 50\,Hz force-control policy reuses the latest token while continuously updating its wrench and proprioceptive histories, enabling reactive interaction regulation without repeatedly running the RFM or compressor.

\begin{figure*}
    \centering
    \includegraphics[width=\linewidth]{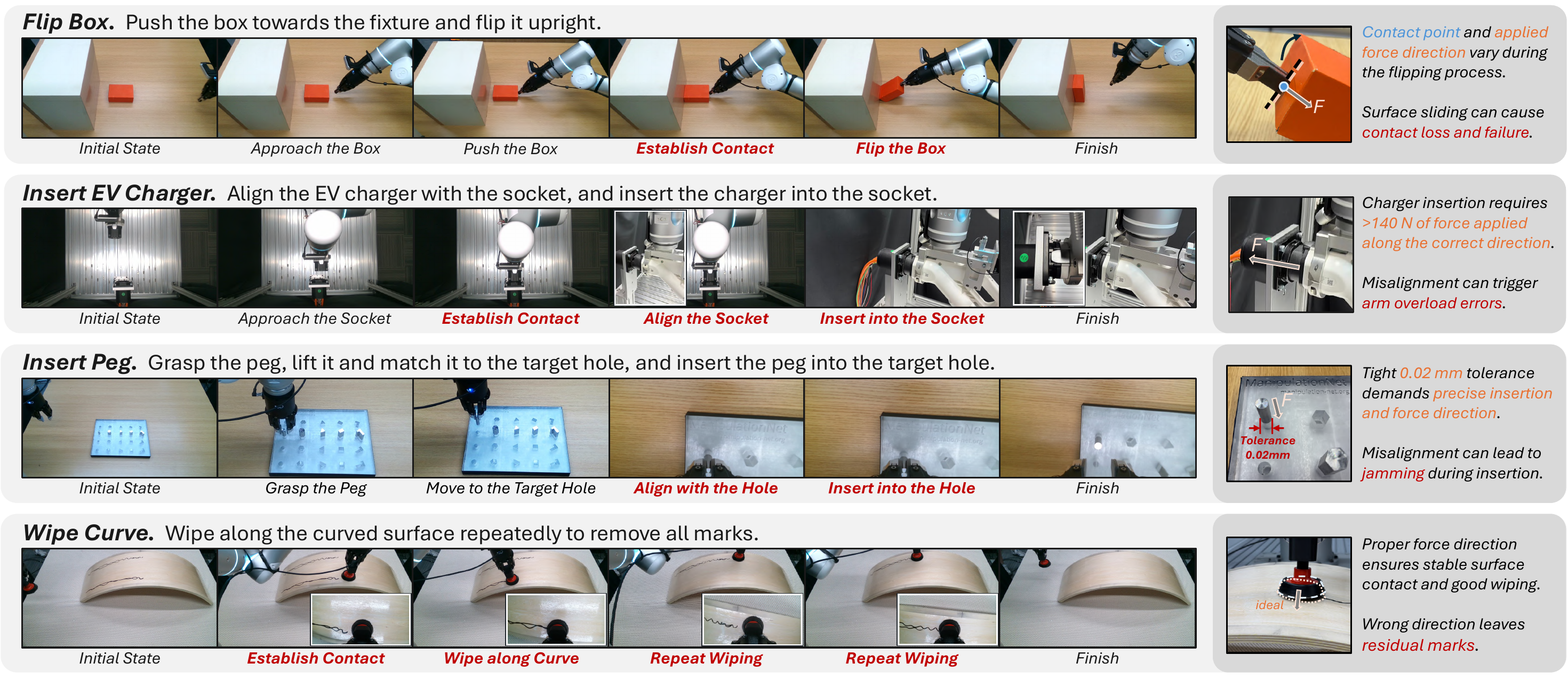}
    \caption{\textbf{Contact-Rich Manipulation Tasks.}
    We evaluate FP2 on 4 real-world tasks covering diverse physical interactions. \textit{\textbf{Flip Box}} requires maintaining effective contact as the contact point and force direction vary during flipping; \textit{\textbf{Insert EV Charger}} requires large insertion forces along the correct direction; \textit{\textbf{Insert Peg}} demands precise alignment under a tight tolerance; and \textit{\textbf{Wipe Curve}} requires sustained contact while following a curved surface. Red labels highlight the contact-rich phases, and the right panels illustrate representative challenges.}
    \label{fig:tasks}\vspace{-0.4cm}
\end{figure*}

\subsection{Force-Control Policy}
\label{sec:policy}

Force Policy~\cite{forcepolicy} adopts a global-local architecture, where a global visuomotor policy provides task context and a local force policy jointly predicts robot actions and force-control parameters using wrist vision and physical feedback. FP2 instead adopts an \emph{action-regulation} decomposition tailored to RFMs: the foundation policy is responsible for task-level action generation, while the high-frequency force-control policy focuses solely on regulating physical interaction. This different division of responsibilities motivates two key changes to the high-frequency force-control policy:
\begin{enumerate}
    \item[(1)] \textbf{Remove action generation.} Rather than re-predicting robot actions, the force-control policy predicts only the parameters needed to regulate the actions generated by the foundation policy.
    \item[(2)] \textbf{Remove the wrist-camera input.} Since visual perception and task-level motion generation remain with the foundation policy, its contextual representation already provides the visual and task context needed by the force-control policy. An additional wrist stream would require a separate image encoder, increasing computation and reliance on wrist-specific visual patterns. We then condition the force-control policy only on the compact foundation-policy context and interaction histories.
\end{enumerate}

Interaction regulation is parameterized by three structured signals: the interaction frame $\boldsymbol{\Sigma}_t$, which defines the local coordinates for decoupling force and motion; the force-position selection mask $\mathbf{S}_t$, which specifies the force- and position-controlled directions; and the desired wrench $\hat{\mathbf{W}}_t$, which specifies the target wrench along the force-controlled directions. Together, these signals provide a compact interface to the hybrid force-position controller without requiring the downstream policy to re-predict RFM actions.

Following~\cite{forcepolicy}, supervision for these signals is recovered directly from demonstrations. The force-control policy is trained on the recovered targets while both the foundation policy and contextual token compressor remain frozen (Fig.~\ref{fig:model} (\textit{bottom}), \textbf{Step 3}). Thus, force information is introduced only when training the lightweight force-control policy and is never used to update the foundation policy or compressor.

\subsection{Robot Controller}
\label{sec:control}

The foundation policy runs at 2\,Hz and predicts action chunks sampled at 15\,Hz. To account for inference latency, we apply waypoint dropout following~\cite{forcepolicy}: candidate starting waypoints consistent with the recent motion are identified, and dynamic time warping~\cite{dtw} is used to select the best-aligned starting point. Earlier waypoints are discarded. We then fit a quintic spline to the aligned action chunk and re-sample it at a unified 50\,Hz, producing a smooth trajectory with continuous position, velocity, and acceleration.

The force-control policy also runs at 50\,Hz and predicts $(\boldsymbol{\Sigma}_t,\mathbf{S}_t,\hat{\mathbf{W}}_t)$ synchronously with the RFM trajectory. At each step, the 50\,Hz action reference and force-control parameters are jointly passed to the hybrid force-position controller. The corresponding low-level control law is executed internally at 1\,kHz~\cite{tang2024partially}. This multi-rate design combines 2\,Hz foundation-policy inference with 50\,Hz interaction regulation and 1\,kHz low-level control.
\section{Experiments}\label{sec:exp}

\begin{table*}[t]
    \centering
    \begin{tabular}{lcrcrcrcrc}
        \toprule
        \multirow{2}{*}{\textbf{Policy}}
        & \multicolumn{2}{c}{\textbf{\textit{Flip Box}}}
        & \multicolumn{2}{c}{\textbf{\textit{Insert EV Charger}}}
        & \multicolumn{2}{c}{\textbf{\textit{Insert Peg}}}
        & \multicolumn{2}{c}{\textbf{\textit{Wipe Curve}}}
        & \textbf{Average} \\ \cmidrule(lr){2-3} \cmidrule(lr){4-5} \cmidrule(lr){6-7} \cmidrule(lr){8-9} \cmidrule(lr){10-10}

        & \textbf{SR} $\uparrow$ & \textbf{NFE} $\downarrow$ & \textbf{SR} $\uparrow$ & \textbf{NFE} $\downarrow$ & \textbf{SR} $\uparrow$ & \textbf{NFE} $\downarrow$ & \textbf{CR} $\uparrow$ & \textbf{NFE} $\downarrow$ & \textbf{Score} $\uparrow$ \\ \midrule

        HybridIL~\cite{forcemimic}
        & \result{40\%}{} & 5.496
        & \result{0\%}{} & N/A
        & \result{0\%}{} & N/A
        & \result{48\%}{} & \textbf{0.276}
        & \result{22\%}{} \\

        ACP~\cite{acp}
        & \result{80\%}{} & 14.355
        & \result{20\%}{} & 5.202
        & \result{52\%}{} & 3.794
        & \result{77\%}{} & 1.528
        & \result{57\%}{} \\

        Force Policy~\cite{forcepolicy}
        & \result{84\%}{} & 1.522
        & \result{68\%}{} & 2.157
        & \result{0\%}{} & N/A
        & \result{49\%}{} & 0.406
        & \result{50\%}{} \\

        \midrule

        GR00T N1.7~\cite{gr00t}
        & \result{12\%}{} & 5.836
        & \result{36\%}{} & 1.559
        & \result{0\%}{} & N/A
        & \result{1\%}{} & N/A
        & \result{12\%}{} \\

        \rowcolor[HTML]{F2F2F2}
        GR00T N1.7 + FP2 \textit{(ours)}
        & \resultgain{76\%}{+64} & 2.076 
        & \resultgain{76\%}{+40} & 1.061
        & \result{0\%}{0} & N/A
        & \resultgain{3\%}{+2} & N/A
        & \resultgain{39\%}{+27} \\
        
        \midrule        
        
        LaWAM~\cite{lawam}
        & \result{32\%}{} & 4.329
        & \result{0\%}{} & N/A
        & \result{4\%}{} & 1.955
        & \result{64\%}{} & 3.024
        & \result{25\%}{} \\

        \rowcolor[HTML]{F2F2F2}
        LaWAM + FP2 \textit{(ours)}
        & \resultgain{48\%}{+16} & 1.968
        & \resultgain{20\%}{+20} & 4.309
        & \resultgain{24\%}{+20} & 1.926
        & \resultgain{65\%}{+1} & 1.994
        & \resultgain{39\%}{+14} \\

        \midrule

        $\pi_0$ (\textit{LoRA}) ~\cite{pi0}
        & \result{44\%}{} & 18.531
        & \result{0\%}{} & N/A
        & \result{12\%}{} & 9.112
        & \result{4\%}{} & 3.085
        & \result{15\%}{} \\

        TA-VLA~\cite{tavla}
        & \resultloss{32\%}{-12} & 8.525
        & \resultgain{72\%}{+72} & 1.968
        & \resultloss{4\%}{-8} & 2.587
        & \resultgain{51\%}{+47} & 1.241
        & \resultgain{40\%}{+25} \\

        ForceVLA~\cite{forcevla}
        & \resultgain{60\%}{+16} & 6.837
        & \result{0\%}{0} & N/A
        & \resultgain{36\%}{+24} & 5.958
        & \resultgain{10\%}{+6} & 1.445
        & \resultgain{27\%}{+12} \\

        \rowcolor[HTML]{F2F2F2}
        $\pi_0$ (\textit{LoRA}) + FP2 \textit{(ours)}
        & \resultgain{68\%}{+24} & 5.278
        & \resultgain{\textbf{92\%}}{+92} & \textbf{0.249} 
        & \resultgain{\textbf{76\%}}{+64} & \textbf{1.287}
        & \resultgain{84\%}{+80} & 0.666
        & \resultgain{\textbf{80\%}}{+65} \\

        \midrule

        $\pi_{0.5}$ (\textit{full})~\cite{pi05}
        & \result{40\%}{} & 3.248
        & \result{0\%}{} & N/A
        & \result{32\%}{} & 5.116
        & \result{97\%}{} & 1.488
        & \result{42\%}{} \\

        \rowcolor[HTML]{F2F2F2}
        $\pi_{0.5}$ (\textit{full}) + FP2 \textit{(ours)}
        & \resultgain{\textbf{92\%}}{+52} &  \textbf{0.899}
        & \resultgain{56\%}{+56} & 0.438 
        & \resultgain{52\%}{+20} & 1.647 
        & \resultgain{\textbf{100\%}}{+3} & 0.495
        & \resultgain{75\%}{+33} \\

        \bottomrule
    \end{tabular}
\caption{
\textbf{Real-World Evaluation of Task Performance and Physical Interaction Quality.}
The average score is the arithmetic mean of the four task metrics (SR and CR). Small numbers denote absolute percentage-point changes relative to the corresponding task-adapted RFM. NFE is reported only for valid successful executions; ``N/A'' indicates that no such execution is available.
}
    \label{tab:result}\vspace{-0.4cm}
\end{table*}

We aim to answer the following research questions:
\textbf{(Q1)} Can FP2 improve contact-rich manipulation performance across different RFMs and tasks?
\textbf{(Q2)} How does FP2 compare with integrating force feedback directly into RFMs?
\textbf{(Q3)} Are RFM context and physical feedback complementary for effective force control?
\textbf{(Q4)} What RFM representation provides the most effective context for force control?
\textbf{(Q5)} Do the FP2-specific redesigns of Force Policy~\cite{forcepolicy}, \textit{i.e.}, removing wrist input and action re-generation, improve performance, generalization, and efficiency in the RFM setting?

\subsection{Setup}

\textbf{Platform.}
Our robot platform consists of a Flexiv Rizon 4 arm equipped with a Flexiv GN-02 gripper and a Flexiv FT-03S 6-DoF force/torque sensor mounted at the robot flange. Visual observations are captured by two Intel RealSense D415 RGB-D cameras, one global and one wrist-mounted.

\textbf{Tasks.}
As shown in Fig.~\ref{fig:tasks}, we evaluate FP2 on four real-world contact-rich manipulation tasks: \textit{\textbf{Flip Box}}, \textit{\textbf{Insert EV Charger}}, \textit{\textbf{Insert Peg}}, and \textit{\textbf{Wipe Curve}}, covering diverse challenges in contact maintenance, force-direction regulation, and precise insertion. Their representative interaction difficulties are illustrated in Fig.~\ref{fig:tasks}. In particular, \textit{\textbf{Insert Peg}} follows the setup from ManipulationNet~\cite{manipulationnet} and uses a circular hole with only 0.02\,mm tolerance, making successful insertion highly sensitive to alignment and force regulation.

\textbf{Data.}
We collect synchronized visuomotor and wrench demonstrations via arm-to-arm teleoperation with force feedback~\cite{tdk}: 50 demonstrations each for \textbf{\textit{Flip Box}}, \textbf{\textit{Wipe Curve}}, and \textbf{\textit{Insert EV Charger}}, and 200 for \textbf{\textit{Insert Peg}}.

\textbf{Baselines and Backbones.}
We evaluate FP2 with four RFMs: GR00T N1.7~\cite{gr00t}, LaWAM~\cite{lawam}, $\pi_0$~\cite{pi0}, and $\pi_{0.5}$~\cite{pi05}, covering both VLA and world-model-based robotic policies. We compare against two complementary groups of baselines.
\textbf{(1)} \emph{Force-aware visuomotor policies with explicit force control}, including HybridIL~\cite{forcemimic}, Adaptive Compliance Policy (ACP)~\cite{acp}, and Force Policy~\cite{forcepolicy}. These task-specific methods provide comparisons against dedicated structured or compliant force-control policies.
\textbf{(2)} \emph{Force-aware VLAs}, including TA-VLA~\cite{tavla} and ForceVLA~\cite{forcevla}, which directly incorporate force observations into the foundation policy. For a controlled comparison, TA-VLA, ForceVLA, and FP2 use the same $\pi_0$ backbone with LoRA adaptation~\cite{lora}, the same task demonstrations, and identical evaluation configurations.

\textbf{Metrics.} We report success rate (SR) for \textit{\textbf{Flip Box}}, \textit{\textbf{Insert EV Charger}}, and \textit{\textbf{Insert Peg}}, and completion rate (CR) for \textit{\textbf{Wipe Curve}}, defined as the fraction of marked regions successfully wiped clean. We additionally introduce \emph{Normalized Force Error (NFE)} to quantify physical interaction quality using the 3D force component.
For each task, we temporally normalize demonstration trajectories and compute the mean $\bar{\mathbf{f}}(\tau)$ and standard deviation $\boldsymbol{\sigma}_{\mathbf{f}}(\tau)$ of the demonstrated force profile. We regard
$\bar{\mathbf{f}}(\tau)\pm\boldsymbol{\sigma}_{\mathbf{f}}(\tau)$
as the valid interaction range and penalize only deviations outside this range. For a \textit{successful} rollout $r$, the element-wise deviation is
\begin{equation}
\mathbf{d}_r(\tau_t)
=
\max\left(
\left|\mathbf{f}_r(\tau_t)-\bar{\mathbf{f}}(\tau_t)\right|
-
\boldsymbol{\sigma}_{\mathbf{f}}(\tau_t),
\mathbf{0}
\right).
\end{equation}
We define NFE as force errors normalized by the demonstrated variability,
\begin{equation}
\mathrm{NFE}_r
=
\sqrt{
\frac{1}{3T}
\sum_{t=1}^{T}
\left\|
\frac{\mathbf{d}_r(\tau_t)}
{\boldsymbol{\sigma}_{\mathbf{f}}(\tau_t) + 10^{-8}}
\right\|_2^2
}.
\end{equation}
Lower NFE indicates smaller departures from the demonstrated force range and better interaction quality. We report NFE over successful trials only; for \textit{\textbf{Wipe Curve}}, we consider only successful continuous-wiping executions.

\textbf{Evaluation.} All policies are deployed on a workstation with an NVIDIA RTX 5090 GPU. Following~\cite{cage}, all methods use the same pre-generated test configurations and workspace, with 25 trials per task unless otherwise specified.

\begin{figure*}
    \centering
    \includegraphics[width=\linewidth]{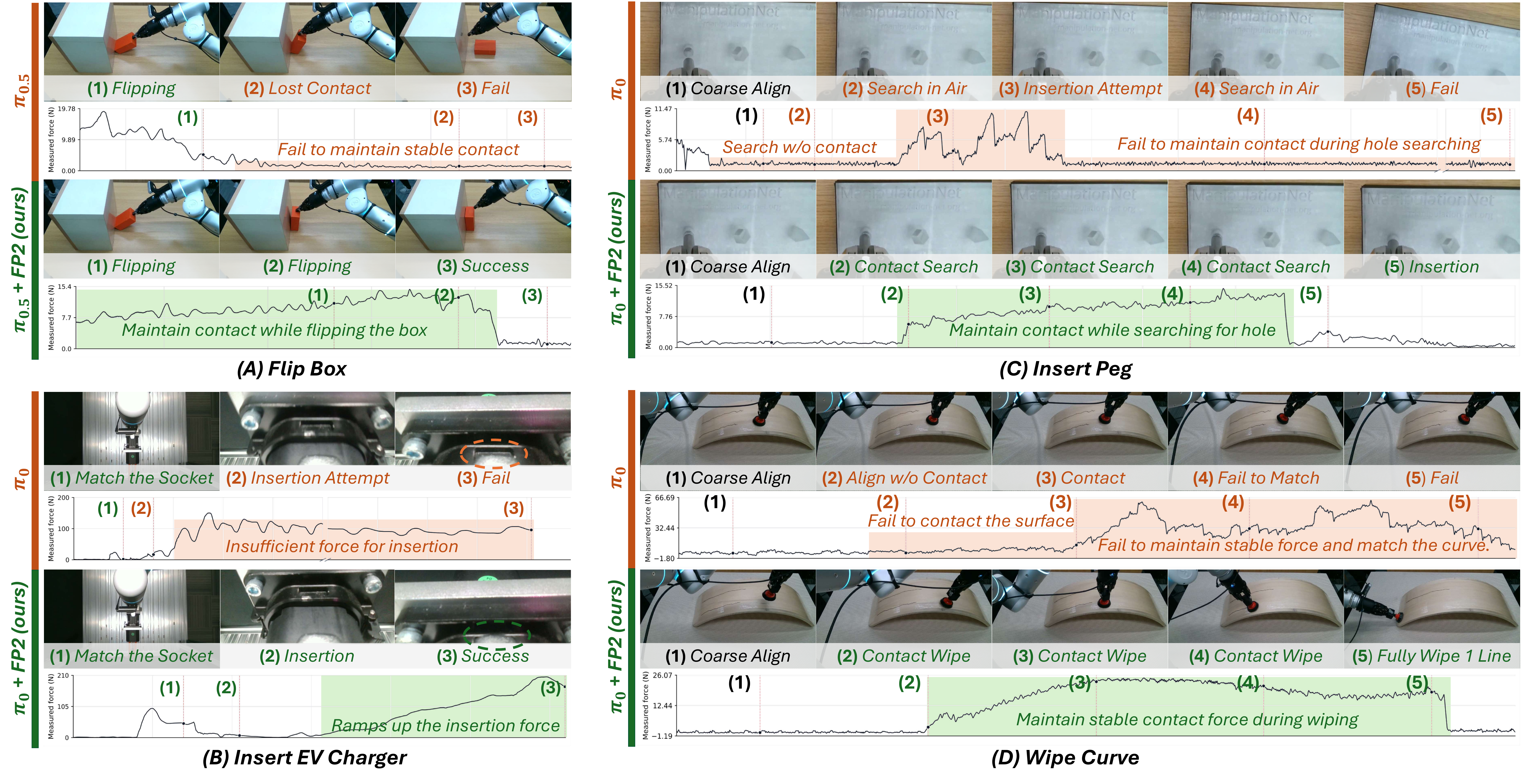}\vspace{-0.2cm}
    \caption{\textbf{Qualitative Analysis of Physical Interaction with and without FP2.}
Representative rollouts and measured force profiles are shown for \textbf{\textit{(A) Flip Box}}, \textbf{\textit{(B) Insert EV Charger}}, \textbf{\textit{(C) Insert Peg}}, and \textbf{\textit{(D) Wipe Curve}}. The foundation policy alone often fails due to lost contact, insufficient force, or unstable interaction, whereas FP2 maintains appropriate force regulation, enabling successful execution.}
    \label{fig:qualitative}\vspace{-0.2cm}
\end{figure*}

\begin{figure*}
    \centering
    \includegraphics[width=\linewidth]{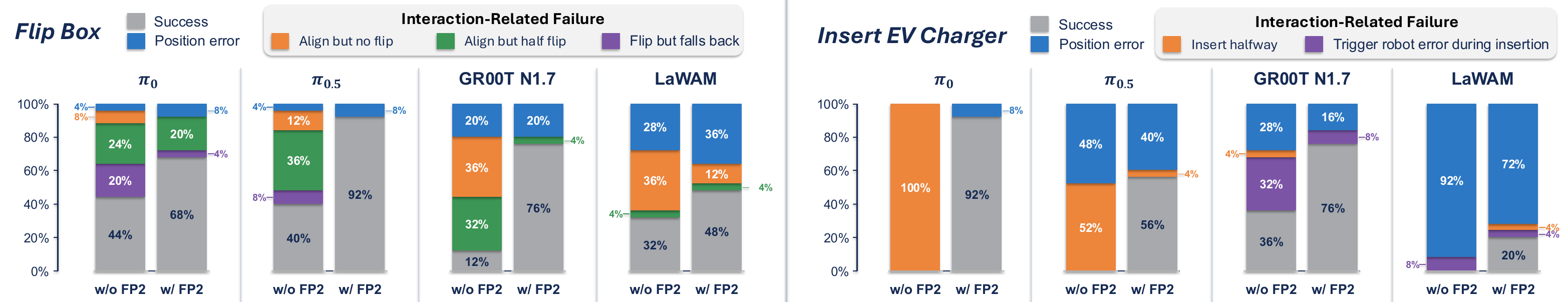}
\caption{
\textbf{Failure Analysis on \textit{Flip Box} and \textit{Insert EV Charger}.}
Each bar summarizes 25 trials and decomposes the outcomes into successful executions and different failure modes, grouped into position errors and interaction-related causes.
}
    \label{fig:failure_analysis}\vspace{-0.4cm}
\end{figure*}

\subsection{Results}

\textbf{FP2 substantially improves contact-rich manipulation across diverse foundation policies and tasks, with gains in both task performance and force regulation quality (Q1).}
As shown in Tab.~\ref{tab:result}, FP2 improves the average performance of all four evaluated RFM backbones and generally reduces NFE relative to the corresponding foundation policy, demonstrating consistent benefits across different policies and interaction patterns. Fig.~\ref{fig:qualitative} further illustrates how this improvement arises during execution: FP2 maintains contact, builds sufficient insertion force, and stabilizes sustained interaction, whereas the foundation policy alone often loses contact or applies inappropriate force. Beyond improving the immediate interaction, maintaining appropriate contact also keeps subsequent robot states closer to the demonstrated interaction regime, mitigating compounding execution errors; this effect is particularly evident in \textit{\textbf{Insert Peg}} and \textit{\textbf{Wipe Curve}}. The gains are naturally smaller when the foundation policy already performs the task well, while failures in task-level motion generation remain beyond the primary scope of FP2. Please refer to \S\ref{sec:failure} for detailed discussions.

\begin{figure*}
\begin{minipage}{0.83\textwidth}
    \footnotesize
    \centering
    \begin{tabular}{ll rr >{\raggedleft\arraybackslash}p{1cm} >{\raggedleft\arraybackslash}p{1cm} >{\raggedleft\arraybackslash}p{1cm} >{\raggedleft\arraybackslash}p{1cm} r}
        \toprule
        \multicolumn{2}{l}{\textbf{Force Control Policy $\pi_F$ Design}} & \multirow{2}{*}{\textbf{Latency}} & \multirow{2}{*}{\textbf{SR}} & \multicolumn{5}{c}{\textbf{SR @ Novel Object}}\\ \cmidrule(lr){1-2} \cmidrule(lr){5-9}
        \textbf{Input} & \textbf{Output} & & & color & texture & stiffness & geometry & \textbf{Average} \\ \midrule
        \multicolumn{9}{l}{\textit{\textbf{(A)} RFM Context and Physical Feedback}}\\
        $\mathbf{z}$ & $(\mathbf{\Sigma},\mathbf{S},\hat{\mathbf{W}})$ & 2.27ms 
        & 4\% & 0/10 & 2/10 & 0/10 & 0/10 & 5\%\\
        
        $(\mathbf{W}, \mathbf{P})$ & $(\mathbf{\Sigma},\mathbf{S},\hat{\mathbf{W}})$ & 4.28ms 
        & 36\% & 4/10 & 2/10 & 1/10 & 0/10 & 18\%\\ 
        
        \midrule    
        \multicolumn{9}{l}{\textit{\textbf{(B)} RFM Context Representation}}\\
        $\text{AvgPool}({\mathbf{C}}),(\mathbf{W},\mathbf{P})$ & $(\mathbf{\Sigma},\mathbf{S},\hat{\mathbf{W}})$ & 4.49ms 
        & 84\% & \textbf{10/10} & 8/10 & 7/10 & 5/10 & 75\% \\
        
        $\text{Latent}(\mathbf{a}), (\mathbf{W},\mathbf{P})$ & $(\mathbf{\Sigma},\mathbf{S},\hat{\mathbf{W}})$ & 4.46ms 
        & 88\% & \textbf{10/10} & \textbf{9/10} & 7/10 & 6/10 & 80\%\\ 
        
        \midrule
        
        \multicolumn{9}{l}{\textit{\textbf{(C)} Wrist Vision and Action Re-generation in Force Control Policy}}\\
        
        $\mathbf{z},  (\mathbf{W},\mathbf{P}), \mathbf{I}_\text{wrist}$ & $(\mathbf{\Sigma},\mathbf{S},\hat{\mathbf{W}})$ & 5.97ms 
        & \textbf{92\%} & 9/10 & 8/10 & 5/10 & 7/10 & 73\%\\ 
        
        $\mathbf{z},  (\mathbf{W},\mathbf{P})$ & $(\mathbf{\Sigma},\mathbf{S},\hat{\mathbf{W}}), \mathbf{a}_F$ & 13.03ms 
        & 76\% & 7/10 & 8/10 & 5/10 & 1/10 & 53\%\\
        
        $\mathbf{z},  (\mathbf{W},\mathbf{P}), \mathbf{I}_\text{wrist}$ & $(\mathbf{\Sigma},\mathbf{S},\hat{\mathbf{W}}), \mathbf{a}_F$ & 14.44ms 
        & 88\% & 9/10 & 9/10 & 6/10 & 5/10 & 73\% \\ \midrule
        
        \multicolumn{9}{l}{\textit{\textbf{Ours:} FP2 Force Control Policy Design}}\\
        \rowcolor[HTML]{F2F2F2}
        $\mathbf{z},(\mathbf{W},\mathbf{P})$
        & $(\mathbf{\Sigma},\mathbf{S},\hat{\mathbf{W}})$
        & 4.48ms & \textbf{92\%}
        & \textbf{10/10} & \textbf{9/10} & \textbf{9/10} & \textbf{7/10}
        & \textbf{88\%} \\
        \bottomrule
    \end{tabular}
\end{minipage}
\hfill
\begin{minipage}{0.16\textwidth}
    \centering
    \includegraphics[height=6.15cm]{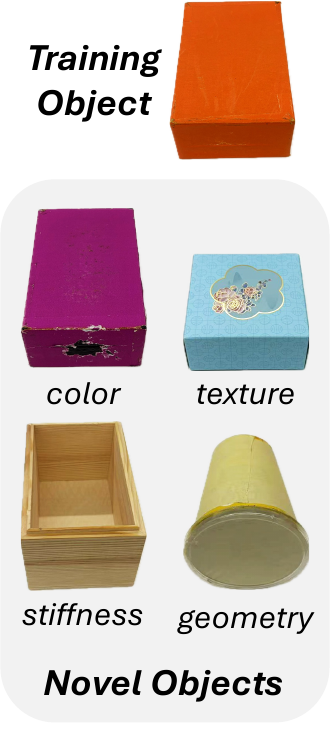}
\end{minipage}
    \captionof{table}{\textbf{Ablation of FP2 Force Control Policy Design on the \textit{Flip Box} Task with $\pi_{0.5}$.} \textit{\textbf{(Left)}} We compare different inputs \textbf{\textit{(A)}}, RFM context representations \textbf{\textit{(B)}}, wrist vision usage \textbf{\textit{(C)}}, and action re-generation \textbf{\textit{(C)}}, and report both training object and novel object success rates. Latency is measured per inference on an RTX 5090 GPU. \textit{\textbf{(Right)}} Illustrations of the training object and novel objects. These novel objects differ from the training object in terms of color, texture, stiffness, and geometry, respectively.}
    \label{tab:ablation}\vspace{-0.4cm}
\end{figure*}

\textbf{FP2 also compares favorably with task-specific force-control policies in both task performance and force regulation quality (Q1).}
The comparison with explicit force-control baselines~\cite{forcepolicy,acp,forcemimic} isolates the benefit of leveraging a foundation policy beyond force control itself. While these task-specific approaches learn manipulation behavior and force regulation jointly, FP2 retains the perception and action capabilities of the foundation policy and learns only downstream interaction regulation. When paired with capable foundation policies such as the $\pi$-series, FP2 achieves substantially higher average task performance while maintaining strong force regulation quality. In particular, $\pi$-series + FP2 achieves lower NFE than the explicit force-control baselines on three of the four tasks and remains competitive on \textit{\textbf{Wipe Curve}}. These results highlight the benefit of reusing foundation-policy capabilities while learning only downstream interaction regulation.

\textbf{Downstream force regulation yields more consistent gains than directly integrating force into the RFM (Q2).}
Under the matched $\pi_0$ setting in Tab.~\ref{tab:result}, force-aware VLAs that introduce force into the foundation policy show strongly task-dependent behavior: TA-VLA~\cite{tavla} improves some tasks while degrading others, and ForceVLA~\cite{forcevla} provides less consistent gains. In contrast, FP2 improves $\pi_0$ across all four tasks and achieves a substantially higher average score. To rule out limited LoRA adaptation as the main cause, we further fully fine-tune all methods on the challenging \textit{\textbf{Insert Peg}} task. Full fine-tuning raises vanilla $\pi_0$ from 12\% to 60\%, yet TA-VLA and ForceVLA achieve only 36\% and 32\%, respectively, while FP2 reaches 76\% under both LoRA and full fine-tuning. This suggests that directly introducing force into the RFM is not guaranteed to preserve its task-level behavior, whereas FP2 leaves action generation in the foundation policy and adds force regulation downstream. The separation also enables interaction regulation to run independently at 50\,Hz, faster than the RFM action loop, which is desirable for rapidly evolving contact dynamics~\cite{lag-fusion,forcepolicy}. Overall, the results support preserving task-level action generation in the foundation policy while introducing force through a dedicated downstream regulation policy.

\subsection{Failure Analysis}\label{sec:failure}
\textbf{Failure analysis further explains where the gains of FP2 come from and where its capability is limited (Q1).}
As shown in Fig.~\ref{fig:failure_analysis}, adding FP2 consistently shifts failures away from interaction-related causes, such as insufficient force and unstable contact. The remaining failures are more often associated with position mismatch and geometric alignment, which lie outside the primary role of the force control policy. This indicates that FP2 is most effective when the foundation policy provides a viable task-level motion: it stabilizes its physical execution and suppresses interaction-induced deviations, but does not replace the foundation policy for task-level action generation or geometric reasoning.

\subsection{Ablations}\label{sec:ablation}

\textbf{RFM context and physical feedback are complementary, and both are essential for effective force control (Q3).}
As shown in Tab.~\ref{tab:ablation}(A), using either the compressed RFM context $\mathbf{z}$ or proprioceptive histories $(\mathbf{W},\mathbf{P})$ alone leads to poor performance, while combining them substantially improves both seen- and novel-object success. This confirms that robust force regulation requires both the manipulation intent provided by RFM context and the realized interaction captured by physical feedback.

\textbf{The compressed contextual token provides the most effective RFM representation for conditioning force control (Q4).}
As shown in Tab.~\ref{tab:ablation}(B), both average-pooled contextual tokens and action-expert latents provide useful conditioning, while the learned compressed token achieves the best performance with comparable latency. This suggests that RFM contextual tokens retain richer manipulation context than the action-expert latent, and that the learned bottleneck preserves this information more effectively than simple token aggregation~\cite{huang2026unireflex} for downstream force control.

\textbf{Removing wrist vision and decoupling action generation from the force-control policy $\pi_F$ improve both generalization and efficiency (Q5).}
As shown in Tab.~\ref{tab:ablation}(C), adding wrist observations to $\pi_F$ provides no benefit on seen objects, while reducing novel-object generalization and increasing inference latency. More importantly, allowing $\pi_F$ to re-generate actions~\cite{huang2026unireflex} substantially degrades both seen- and novel-object performance and nearly triples inference latency. Joint action and force regulation prediction with wrist vision~\cite{forcepolicy} also remains inferior to FP2. These results support the action-regulation decomposition of FP2: the foundation policy retains responsibility for action generation, while the lightweight force-control policy focuses solely on physical interaction regulation.

\section{Conclusion}\label{sec:conclusion}

Our results show that effective force control for robotic foundation models benefits from separating \emph{action generation} from \emph{interaction regulation}. FP2 retains task-level action generation in the foundation policy and introduces a lightweight, high-frequency force control policy that conditions on foundation-policy context and physical feedback to regulate interaction. Across multiple foundation policies and diverse contact-rich tasks, this decomposition consistently improves task performance and force regulation quality. These results highlight a practical path for equipping existing RFMs with force control without modifying their native sensing or action-generation interface.

\textbf{Limitations and Future Work.}
FP2 assumes that the foundation policy provides a viable task-level motion and cannot fully recover from incorrect action generation or geometric reasoning. Its force regulation is also currently learned at the task level; a shared multi-task force control policy could enable reusable interaction skills across tasks and embodiments. More broadly, future RFMs could support a bidirectional interface with force control, providing interaction intent for downstream regulation while using summarized physical feedback to adapt subsequent actions.

\section*{Acknowledgement}

We would like to thank Wenbo Tang at Flexiv for tuning the non-real-time hybrid force-position controller, and Zhipeng Zhang at Flexiv for tuning the arm-to-arm teleoperation system. We also thank Jinghong Liu, Ting Yin, and Yi Wu at Flexiv, and Dong Liu, Yuanyuan Jia at Noematrix, for designing and printing the fixtures used in our experiments.  We thank Shangning Xia at Noematrix for valuable discussions, Peishen Yan at Shanghai Jiao Tong University for proofreading, and Junchao Zhang at Noematrix for his assistance in data collection.

\printbibliography

\end{document}